\def\paperlanguage{}
\pdfoutput=1 

\documentclass[letterpaper, 10 pt, conference]{ieeeconf}  

\usepackage{bm}
\usepackage{cite}
\usepackage{flushend}
\include{preamble}

\IEEEoverridecommandlockouts                              

\title{\LARGE \textbf
  {
    \switchlanguage%
    {%
      Online Learning of Joint-Muscle Mapping Using Vision in Tendon-driven Musculoskeletal Humanoids
    }%
    {%
      筋骨格ヒューマノイドにおける\\視覚を利用した関節-筋空間マップの逐次的再学習
    }%
  }
}

\author{Kento Kawaharazuka, Shogo Makino, Masaya Kawamura, Yuki Asano, Kei Okada and Masayuki Inaba
  \thanks{Authors are with Department of Mechano-Informatics, Graduate School of Information Science and Technology, The University of Tokyo, 7-3-1 Hongo, Bunkyo-ku, Tokyo, 113-8656, Japan.
    {\texttt\small [kawaharazuka, makino, kawamura, asano, k-okada, inaba]@jsk.t.u-tokyo.ac.jp}
  }
}
\begin{document}

\maketitle
\thispagestyle{empty}
\pagestyle{empty}

\begin{abstract}
  \switchlanguage%
  {%
    The body structures of tendon-driven musculoskeletal humanoids are complex, and accurate modeling is difficult, because they are made by imitating the body structures of human beings.
    For this reason, we have not been able to move them accurately like ordinary humanoids driven by actuators in each axis, and large internal muscle tension and slack of tendon wires have emerged by the model error between its geometric model and the actual robot.
    Therefore, we construct a joint-muscle mapping (JMM) using a neural network (NN), which expresses a nonlinear relationship between joint angles and muscle lengths, and aim to move tendon-driven musculoskeletal humanoids accurately by updating the JMM online from data of the actual robot.
    In this study, the JMM is updated online by using the vision of the robot so that it moves to the correct position (Vision Updater).
    Also, we execute another update to modify muscle antagonisms correctly (Antagonism Updater).
    By using these two updaters, the error between the target and actual joint angles decrease to about 40\% in 5 minutes, and we show through a manipulation experiment that the tendon-driven musculoskeletal humanoid Kengoro becomes able to move as intended.
    This novel system can adapt to the state change and growth of robots, because it updates the JMM online successively.
  }%
  {%
    筋骨格ヒューマノイドは人間を模倣しているが故に身体が複雑であり、正しいモデル化が難しい。
    そのため、通常のヒューマノイドのように指令通りに体を動かすのは難しく、また、実機とモデルの間の誤差によって大きな内力や筋の緩みが発生していた。
    そこで我々は、関節角と筋長の非線形な関係を表す関節-筋空間マッピングをニューラルネットワークとして表現し、それを実機の動作データからオンラインでアップデートしていくことで体を指令通りに動かせるようになることを目指した。
    本研究では視覚を利用し、関節-筋空間マッピングを正しい方向にアップデートしていく。
    また、内力や筋の緩みを修正していくようなアップデートも行う。
    以上二つのオンライン学習によって、筋骨格ヒューマノイド腱悟郎が正しい動作を行うことができるようになることを物体把持実験を通して示す。
    またこの手法は、オンラインで逐次的に更新を行っていくため、ロボットの状態変化や成長にも対応することができる。
  }%
\end{abstract}

\section{INTRODUCTION} \label{sec:1}
\begin{figure}[t]
  \centering
  \includegraphics[width=1.0\columnwidth]{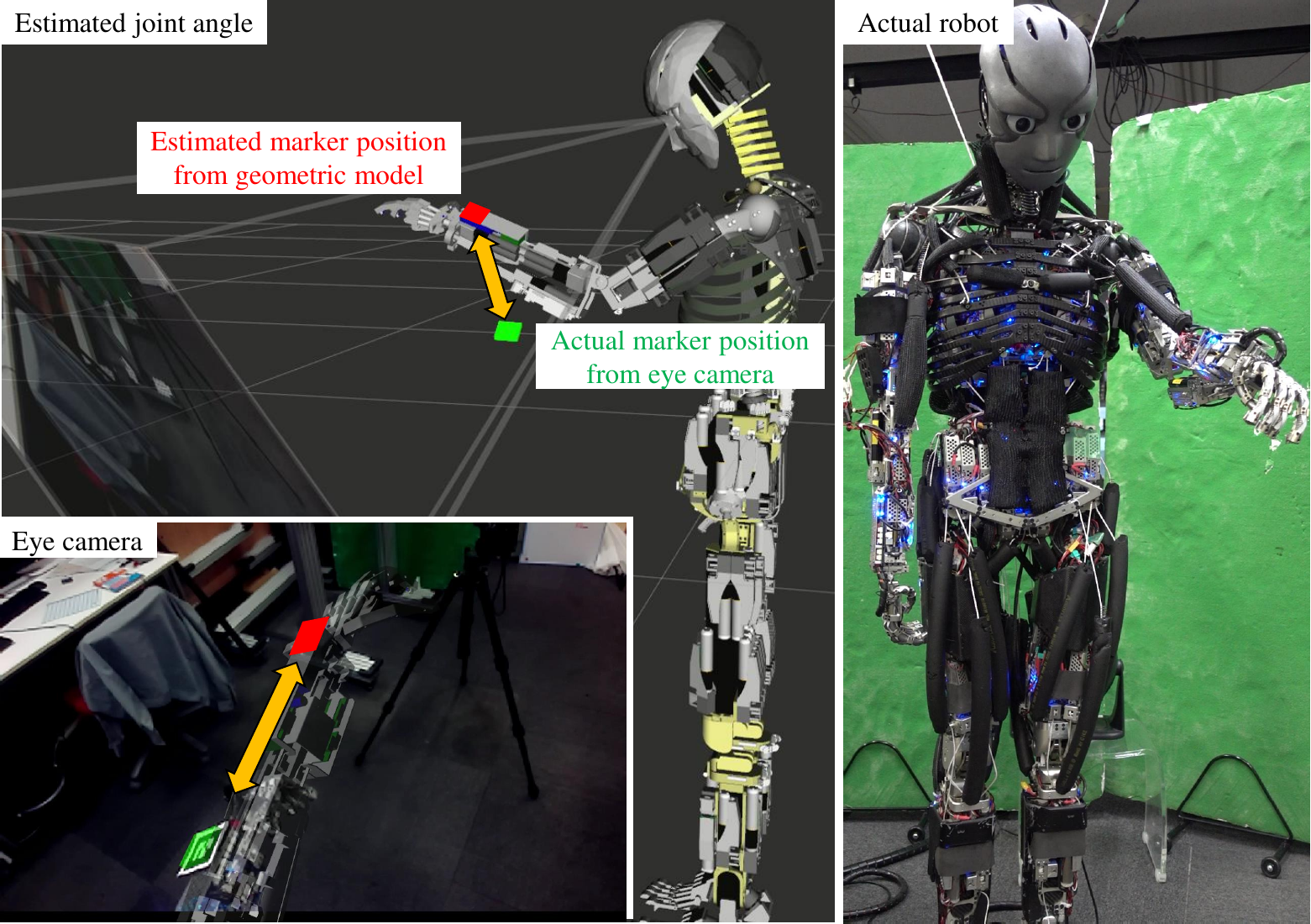}
  \vspace{-3.0ex}
  \caption{Difficulty of movement to the intended position. This figure shows the difference between the movement of a geometric model and the movement of the actual robot when grasping object.}
  \label{figure:intended-move-difficulty}
  \vspace{-1.0ex}
\end{figure}

\switchlanguage%
{%
  Tendon-driven musculoskeletal humanoids \cite{ijars2013:nakanishi:approach, artl2013:wittmeier:ecce, humanoids2013:michael:anthrob, humanoids2016:asano:kengoro} are expected to play an active part in human society in the future, because they have multiple degrees of freedom (multi-DOFs) like the scapula and spine of human beings, can realize variable stiffness, are soft in contact, etc.
  However, the accurate modeling of tendon-driven musculoskeletal humanoids is difficult, because they are made by imitating human beings and so they have complex muscle arrangements and body structures.
  It is difficult to move them accurately like ordinary humanoids driven by actuators in each axis, if we use only their geometric models (\figref{figure:intended-move-difficulty}).
  Also, internal muscle tension and slack of tendon wires tend to emerge because of the model error between its geometric model and the actual robot.
  Thus, in order for tendon-driven musculoskeletal humanoids to gain popularity, we need a novel system by which an exact geometric model is not necessary, and instead the movement information of the actual robot is used in order to move the robots accurately as intended.

  In previous studies, various controls were developed in order to move tendon-driven musculoskeletal humanoids as intended, and decrease the internal muscle tension and slack of tendon wires.
  Ookubo, et al. \cite{humanoids2015:okubo:muscle-learning} and J{\"a}ntsch, et al. \cite{iros2012:michael:elbow} obtained data of joint angles and muscle lengths from the actual robot by moving the humanoid, acquired a joint-muscle mapping (JMM) which expresses a nonlinear relationship between joint angles and muscle lengths by polynomial regression of the data, and moved it accurately by using this JMM.
  Also, Kawaharazuka, et al. \cite{ral2017:kawaharazuka:aic} estimated the muscle antagonism and decreased the internal muscle tension and slack of tendon wires by inhibiting antagonist muscles against agonist muscles.
  However, these methods do not update JMM online, and cannot adapt to the state change of the actual robots according to the elongation of tendon wires.
  Also, in the studies of \cite{humanoids2015:okubo:muscle-learning} and \cite{iros2012:michael:elbow}, we must obtain numerous data from the actual robot, and it is difficult to apply these methods to JMMs including multi-DOFs like human beings due to its computational complexity.
  In other studies, several controls consider muscle synergy \cite{fcn2013:allessandro:synergy}.
  Among them, Diamond et al. \cite{biobio2014:diamond:reaching} implemented an algorithm using muscle synergy and reinforcement learning for a reaching task.
  This method realizes a simple algorithm using muscle synergy, and the reaching movement can be improved in each trial, but the experiment is done only in a simulation environment, and there are some problems regarding the number of trials, lack of versatility, etc.
}%
{%
  筋骨格ヒューマノイド\cite{humanoids2016:asano:kengoro, humanoids2010:hugo:eccerobot}は人間のような多自由度や可変剛性の実現, やわらかい接触などにおいて優れており, 今後の活躍が期待される.
  しかし, 人間を模した複雑な筋配置や骨格構造をしているがゆえに正しいモデル化が難しい。
  ゆえに、通常のヒューマノイドのように指令通りの場所に自身の体を動かすのは困難である(\figref{figure:intended-move-difficulty}).
  また, モデルと実機の間の誤差によって筋肉の弛みや内力の高まりが生じやすい.
  そこで, 筋骨格ヒューマノイドを普及させるためには, 正しい幾何モデルを必要とせず, 指令通りに自身の関節を動かせるシステムが必要となる.

  過去の研究では、指令通りの動作を行うため、そして、内力や筋の緩みを減らすために様々な制御が行われてきた。
  \cite{humanoids2015:okubo:muscle-learning}や\cite{iros2012:michael:elbow}では最初に体を動作させ、その際に実機から得たデータから関節-筋空間マッピング多項式近似によって求めることで指令通りの動作を行えるように工夫している。
  \cite{ral2017:kawaharazuka:aic}では拮抗関係を推定し、主動筋に対して拮抗筋を抑制することで内力の高まりや筋の緩みを抑えている。
  しかしこれらは、オンラインで一切の学習をしておらず、ロボットの状態変化に対応することができない。
  また、最初の二つは初めに人間が大量のデータセットを作らなければならず、計算量・時間的に人間の多自由度な体全体に置いて関節-筋空間マッピングを求めることは難しい。
  他にも、\cite{fcn2013:allessandro:synergy}ではmuscle synergyを考慮した様々な制御方法が紹介されている。
  その中でも\cite{biobio2014:diamond:reaching}はmuscle synergiesの考え方と強化学習によってリーチング動作を行うアルゴリズムを実装している。
  この方法は、muscle synergiesを使うことでsimpleなアルゴリズムを実現しており、施行するごとに動作が改善することが見込まれるが、シミュレーションのみであり、試行回数や汎用性の観点から問題がある。
}%

\switchlanguage%
{%
  To solve these problems, we propose a novel system which conducts online learning of JMM using the vision of tendon-driven musculoskeletal humanoids.
  In this system, JMM is expressed using neural network (NN), and muscle jacobian is obtained by differentiating the JMM.
  The humanoid itself looks at part of its body such as its hands and feet, and becomes able to move accurately by modifying the JMM online using data of the actual robot.
  At the same time, we do another online update of the JMM to correct internal muscle tension and slack of tendon wires.
  This study does not consider the influence of muscle compliance, and deals with only the condition in which no forces, except for gravity, are applied to the robot.
  Also, we assume that the geometric model regarding the joint structure of the robot is correct.

  Using this proposed system, manipulation tasks, which were challenging for tendon-driven musculoskeletal humanoids to execute using only its geometric model, can be done by updating JMM accurately online using the data of the actual robot.
  Thus, this system can be adapted to the growth and state change of robots.

  This paper is organized as follows.
  In \secref{sec:1}, we stated the motivation and the goal of this study.
  In \secref{sec:2}, as an introduction to this system, we will explain the method of training JMM, obtaining muscle jacobian, updating JMM online, and estimating joint angles.
  In \secref{sec:3}, we will explain the overview of this system and two types of online learnings of JMM.
  In \secref{sec:4}, we will conduct some basic experiments about respective components of this system, and discuss the effectiveness of this study through a manipulation task experiment.
  Finally, in \secref{sec:5}, we will state the conclusion and the future works.
}%
{%
  そのために我々は, 視覚を用いた関節-筋空間マップの逐次的再学習を行うシステムを提案する.
  このシステムにおいて, 関節-筋空間マップはニューラルネットワーク(以下NN)によって表現され, それを微分することによって筋長ヤコビアンを得る.
  そして, ロボット自身が自分の手や足など, 体の一部分を見ることによって関節-筋空間マップをオンラインで修正し, より指令した通りに体を動かせるようになる.
  また同時に, 筋の緩みや内力の高まりを修正していくようなオンラインのNNの修正も行う.
  ただし、本研究は筋のコンプライアンスの影響は考慮していなく、重力以外の外力が加わっていない状態にのみ適用可能である。
  また、ロボットのリンク構造に関する幾何モデルは既知とする。
  本研究で提案するシステムを用いれば, 幾何モデル情報からだけでは正しくアプローチ出来なかった物体把持を, 試行動作によって関節-筋空間マップを修正して成功することができ, ロボットの成長や状態変化にも逐次的に対応することができるようになる.

  本研究は以下のような構成となっている。
  \secref{sec:1}では本論文の動機と目標について述べた。
  \secref{sec:2}では前準備として、ニューラルネットワークによる関節-筋空間マップの学習方法、筋長ヤコビアンの導出方法、オンライン学習方法、関節角度推定方法について述べる。
  \secref{sec:3}では本研究のシステム全体、そして二つのオンライン学習について説明する。
  \secref{sec:4}ではこのシステムの個々の基本的な実験を行い、最後に、物体把持に関する応用実験を行うことで有効性を確かめる。
  最後に、\secref{sec:5}では結論と今後の方針について述べる。
}%

\section{Learning of Joint-Muscle Mapping using a Neural Network} \label{sec:2}
\begin{figure}[htb]
  \centering
  \includegraphics[width=1.0\columnwidth]{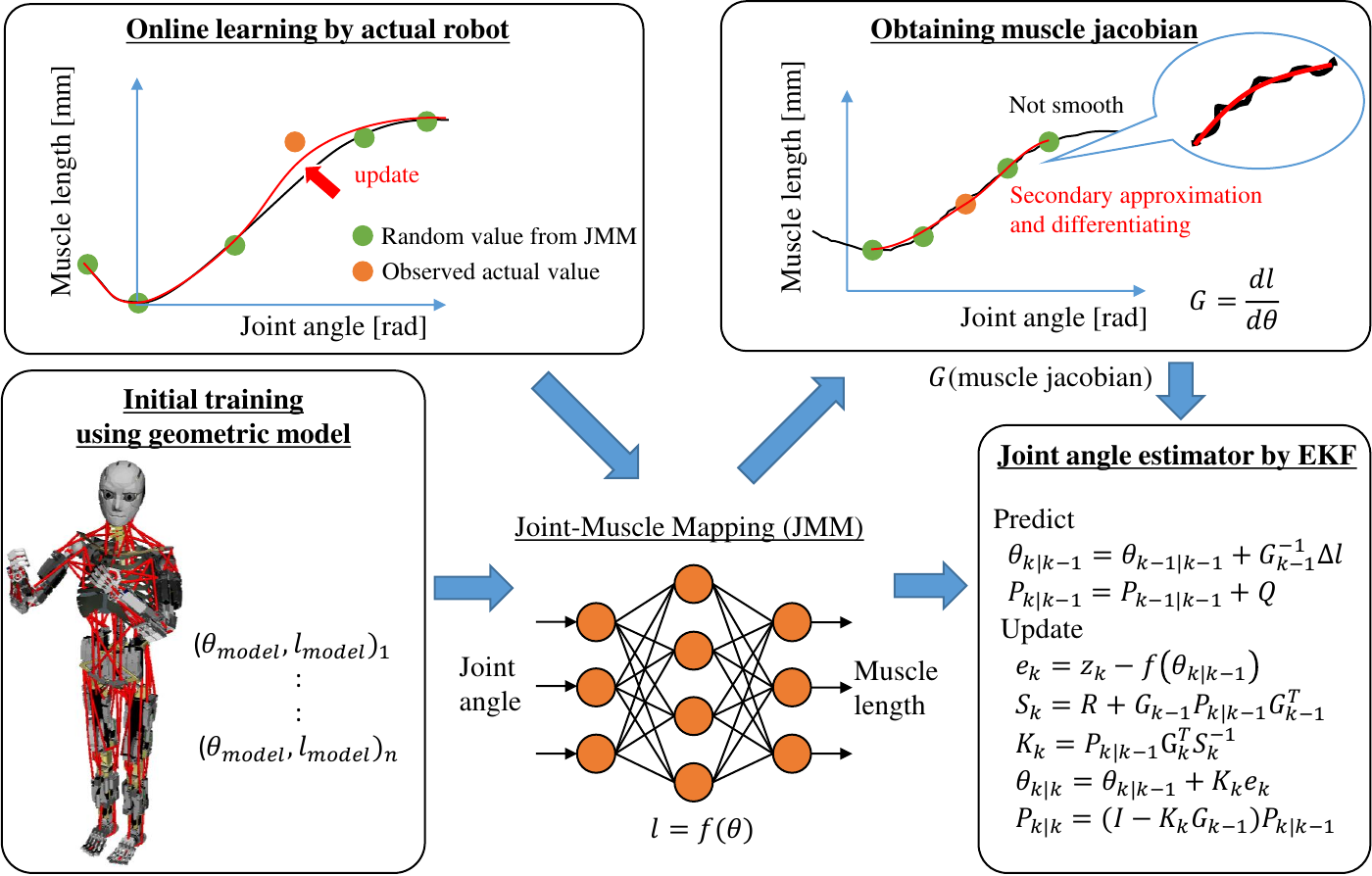}
  \vspace{-3.0ex}
  \caption{Overview of the expression of joint-muscle mapping using neural network.}
  \label{figure:neural-network-system}
  \vspace{-1.0ex}
\end{figure}

\switchlanguage%
{%
  To express JMM, methods using table-searching \cite{icra2010:nakanishi:table}, polynomial regression \cite{iros2012:michael:elbow, humanoids2015:okubo:muscle-learning}, and NN \cite{robio2011:michael:control} have been proposed.
  The methods using table-searching and polynomial regression have a problem that the computational cost increases exponentially in accordance with the increase of included joints, and are not good at online updating of JMM.
  The positive side to using polynomial regression is that we can obtain smooth muscle jacobian by differentiating.
  In comparison, the expression using NN is good at online updating, but we cannot obtain smooth muscle jacobian by differentiating.
  J{\"a}ntsch, et al. \cite{robio2011:michael:control} collected the data set to train NN only at first, and there is no discussion about the methods to obtain smooth muscle jacobian, to estimate joint angles, and to update JMM online.

  Thus, in this study, we express JMM using NN in order to make online learning of JMM possible.
  At first, though the difference between the geometric model and the actual robot is large, we construct JMM using the information of the geometric model.
  After that, we modify the JMM obtained from the geometric model using the information from the movement of the actual robot.
  In this section, we will explain the method of training JMM from a geometric model, obtaining smoothened muscle jacobian, updating JMM online, and estimating joint angles.
  The overview of the expression of JMM using NN is shown in \figref{figure:neural-network-system}, and we will explain the respective components below.
}%
{%
  関節-筋空間マップの表現としてはテーブル探索による方法\cite{icra2010:nakanishi:table}や多項式近似による方法\cite{humanoids2015:okubo:muscle-learning}, NNによる方法\cite{robio2011:michael:control}が提案されている.
  この中でも, テーブル探索による方法と多項式近似による方法は関節数の増加に対して計算量が指数的に増大し, また, オンラインでのマップの更新にも向いていない.
  多項式近似の良いところは、微分によって滑らかな筋長ヤコビアンが得られることである。
  それに対して, NNによる表現ではバッチ処理ではないため再学習に適しているが、滑らかの筋長ヤコビアンは得られない。
  \cite{robio2011:michael:control}では最初にデータセットを集め学習させることのみを行っており, 滑らかな筋長ヤコビアンの導出のための工夫や関節角度の推定, オンラインでの再学習などに関する議論はなされていない.

  本研究ではまず、関節-筋空間マップ, つまり, 関節角度と筋長の間の非線形な関係を, オンラインでの再学習に対応できるようにNNを用いて表す.
  実機とのズレは大きいが、最初は簡単な幾何モデルからの情報を用いて、関節-筋空間マッピングを構築する.
  その後、実機からのデータを用いて幾何モデルから得た関節-筋空間マッピングを正しいものへと修正していく。
  本章ではシステム全体の構成に入る前に, 関節-筋空間マップをNNでどのように学習し, 微分による筋長ヤコビアンの導出, 関節角度の推定, オンラインでの再学習を行うのかを説明する.
  このシステムの概要は\figref{figure:neural-network-system}のようになっており, それぞれのコンポーネントについて以下で説明する.
}%

\subsection{Initial Training of Joint-Muscle Mapping from a Geometric Model} \label{subsec:initial-learning}
\switchlanguage%
{%
  In this study, a function which expresses a nonlinear relationship between joint angles and muscle lengths:
  \begin{align}
    \bm{l} = f(\bm{\theta})
  \end{align}
  is expressed by neural network, which has 3 layers: an input layer of joint angles, a hidden layer, and an output layer of muscle lengths.
  $\bm{l}$ is muscle lengths and $\bm{\theta}$ is joint angles.

  First, we construct a simple geometric model of a tendon-driven musculoskeletal humanoid.
  In this study, we use Kengoro \cite{humanoids2016:asano:kengoro} (the details are in \secref{subsec:basic-experiments}).
  This geometric model is a simple one composed of muscles in which the starting point, relay points, and end point are connected (the lower left figure of \figref{figure:neural-network-system}).
  Next, we construct JMM for each respective body part such as the shoulder, scapula, neck, and so on.
  For the example of the shoulder, when we construct the JMM including 3 DOFs glenohumeral joint and 1 DOF elbow joint, we pick up muscles which move these 4 DOFs (in this study, 10).
  If these muscles include polyarticular muscles which move other DOFs beside these 4 DOFs, we must add those other DOFs to the JMM of the shoulder.
  Thus, the JMM of 4 DOFs around the shoulder is the nonlinear relationship between 10 muscles which move these 4 DOFs and all joints which these 10 muscles can move (for example, the pectoralis major muscle can move scapula joints in addition to the shoulder joints).
  In this case, there are 8 DOFs including the 4 DOFs scapula (we limit the DOF of the scapula to 4 DOFs: roll and yaw of sternoclavicular joint, and roll and pitch of acromioclavicular joint).
  Through this process, the JMM of the shoulder is composed of 8 DOFs and 10 muscles.
  We construct the other JMMs likewise.
  Although it is possible to construct a JMM including all DOFs and muscles of the entire body, this is difficult due to its computational complexity, and this is one of our future works.

  Finally, we train these JMMs by the geometric model of the tendon-driven musculoskeletal humanoid.
  We move respective joint angles of the geometric model little by little, calculate the muscle lengths, and construct a data set.
  Using this data set, we train NN of JMM.

  The JMM obtained from the geometric model has almost the same value as the geometric model.
  However, as shown in \figref{figure:intended-move-difficulty}, there is a large error between this JMM and the actual robot, so we need to modify it using the information from the movement of the actual robot.
}%
{%
  関節角度と筋長の間の非線形な関係を表す関数
  \begin{align}
    \bm{l} = f(\bm{\theta})
  \end{align}
  を構成するNNは, 入力層である関節角度, 隠れ層, 出力層である筋長の3層で表されたNNを用いる.
  ここで$\bm{l}$は筋長、$\bm{\theta}$は関節角度を表す。

  まず, 筋骨格ヒューマノイドの簡易的な幾何モデルを作成する.
  本研究では筋骨格ヒューマノイド腱悟郎\cite{humanoids2016:asano:kengoro}を用いる.
  この幾何モデルは筋肉の起始点・中継点・終止点を直線で結んだ簡単なモデルである (\figref{figure:neural-network-system}の左下図).

  次に, 関節-筋空間マップを身体部位ごとに分割する.
  例えば肩周辺であれば肩の3軸と肘の1軸に関する関節-筋空間マップを形成するが, その4軸を動作させる筋を抽出し(本研究では10本), さらにそれらの筋が, 肩周辺の4軸以外の関節も動作させる多関節筋であればその関節もマップに加える.
  つまり肩関節周辺の4軸に関する関節-筋空間マップは, それらを動作させる10本の筋と, それら10本の筋が動作させる関節すべて(大胸筋などは肩甲骨関節も動作させる), つまり肩甲骨の4軸(本研究では肩甲骨を胸鎖関節のroll, yawと肩鎖関節のroll, pitchに限定しているので4軸)を含めた8軸との非線形な関係を示すものとなる.
  よって8軸と10本の筋に関するNNが構成された.
  その他の関節群に関しても同様である.
  体全体の自由度と筋を対応付ける関節-筋空間マッピングを作成しても良いが、計算量的観点から今後の課題とする。

  最後に, 筋骨格ヒューマノイドの幾何モデルから関節-筋空間マップを学習させる.
  幾何モデルから関節角度を少しずつ動かした際の筋長を算出し, データを作成する.
  それらをNNに入れ, 学習をさせる.

  ここで得た関節-筋空間マップは幾何モデルから得たものであり, 幾何モデルとほとんど等しい値を取る.
  しかし, \figref{figure:intended-move-difficulty}からわかるようにこの関節筋空間マッピングは実機との間に大きな誤差があり, それを修正していくことが求められる.
}%

\subsection{Derivation of Muscle Jacobian from Neural Network} \label{subsec:jacobian}
\switchlanguage%
{%
  We obtained JMM by training NN using the data set of the geometric model.
  An important component in controlling tendon-driven musculoskeletal humanoids is the muscle jacobian:
  \begin{align}
    G(\bm{\theta}) = d\bm{l}/d\bm{\theta} = df(\bm{\theta})/d\bm{\theta}
  \end{align}
  and we must differentiate NN to obtain this muscle jacobian.
  However, though the JMM expressed using NN is differentiable, because the JMM is the superposition of activation functions, differentiation results in the appearance of each function's features.
  For example, when we use sigmoid as an activation function, the differentiated value can be wavy due to the feature of sigmoid, and when we use ReLu, the value can be jagged in the same way.
  To solve this problem, we obtain muscle jacobian by secondary approximation using some sample points.
  When joint angles of the robot is $\bm{\theta}$, to obtain the muscle jacobian $G_{ij}$ of a certain muscle $i$ and joint $j$, we get some sample points of muscle $i$ length when moving joint $j$ by degrees (in this study, we sampled 5 points: $\theta_j$ and 4 points which have $d_1=10, d_2=20$[deg] interval each from $\theta_j$ in the positive and negative direction), solve secondary approximation from the data set, and obtain smoothened muscle jacobian by differentiating it.
  \begin{align}
    \bm{x}_j = \{\cdots, \theta_j-d_2, \theta_j-d_1, \theta_j, \theta_j+d_1, \theta_j+d_2, \cdots\} \\
    \bm{y}_j = \{\cdots,
    f \begin{pmatrix} \theta_0 \\ \vdots \\ \theta_j-d_1 \\ \vdots \\ \end{pmatrix},
    f \begin{pmatrix} \theta_0 \\ \vdots \\ \theta_j \\ \vdots \\ \end{pmatrix},
    f \begin{pmatrix} \theta_0 \\ \vdots \\ \theta_j+d_1 \\ \vdots \\ \end{pmatrix},
    \cdots\} \\
    \begin{pmatrix}
      \Sigma x_j^4 & \Sigma x_j^3 & \Sigma x_j^2\\
      \Sigma x_j^3 & \Sigma x_j^2 & \Sigma x_j\\
      \Sigma x_j^2 & \Sigma x_j & \Sigma 1\\
    \end{pmatrix}
    \begin{pmatrix}
      a_{ij}\\
      b_{ij}\\
      c_{ij}
    \end{pmatrix}
    =
    \begin{pmatrix}
      \Sigma x_j^2 y_{ij}\\
      \Sigma x_j y_{ij}\\
      \Sigma y_{ij}
    \end{pmatrix} \\
    G_{ij} = 2 a_{ij} \theta_j + b_{ij}
  \end{align}
  where $a_{ij}, b_{ij}, c_{ij}$ are coefficients of secondary approximation, and $y_{ij}$ is the muscle $i$ length of $y_{j}$.

  In addition, there are several other methods to smoothen muscle jacobian, such as the weight decay and minimization of network structure.
}%
{%
  NNの学習によって関節-筋空間の非線形なマップを得ることはできた.
  そして, 筋骨格ヒューマノイドを制御する上で重要であるのが筋長ヤコビアン
  \begin{align}
    G(\bm{\theta}) = d\bm{l}/d\bm{\theta} = df(\bm{\theta})/d\bm{\theta}
  \end{align}
  であるが, これを得るためにはこのマップを微分しなければならない.
  しかし, NNから得たマップは微分可能であるものの、非線形な関数の重ね合わせであるため、微分をすることでその関数自体の性質が現れてしまう.
  活性化関数としてsigmoidを使った場合は微分をすることで重ね合わされたsigmoid関数の性質が現れて微分値が波打ってしまい、ReLuを使った場合は値がギザギザになってしまう。
  そこで本研究では, いくつかのサンプル点を基に二次関数近似を行って筋長ヤコビアンを求める.
  ある姿勢$\bm{\theta}$での, ある筋$i$の, ある関節方向$j$に対する筋長ヤコビアン$G_{ij}$を求めるために, 関節$j$を動かした際の筋$i$の長さのサンプルを数点(本研究では, $\theta_j$を中心に正負に$d=$10[deg]間隔で2点ずつ計5点)取り, 以下のようにそれらを二次曲線で近似してから微分することで滑らかな筋長ヤコビアンの値を得る.
  \begin{align}
    \bm{x}_j = \{.., \theta_j-2d, \theta_j-d, \theta_j, \theta_j+d, \theta_j+2d, ..\} \\
    \begin{pmatrix}
      \Sigma x_j^4 & \Sigma x_j^3 & \Sigma x_j^2\\
      \Sigma x_j^3 & \Sigma x_j^2 & \Sigma x_j\\
      \Sigma x_j^2 & \Sigma x_j & \Sigma 1\\
    \end{pmatrix}
    \begin{pmatrix}
      a_{ij}\\
      b_{ij}\\
      c_{ij}
    \end{pmatrix}
    =
    \begin{pmatrix}
      \Sigma x_j^2 y_{ij}\\
      \Sigma x_j y_{ij}\\
      \Sigma y_{ij}
    \end{pmatrix} \\
    G_{ij} = 2 a_{ij} \theta_j + b_{ij}
  \end{align}
  ここで$a_{ij}, b_{ij}, c_{ij}$は二次関数の係数, $y_{ij}$は, 関節jのみ$\bm{x}_j$で他の関節は元の$\bm{\theta}$であるような関節列を関数$f$に代入したときの筋$i$の筋長列である.
}%

\subsection{Online Learning of Joint-Muscle Mapping from Movement Data of the Actual Robot} \label{subsec:relearn}
\switchlanguage%
{%
  In order to become able to move the actual robot accurately, we need to update the weights of NN obtained from the geometric model.
  However, if we simply update NN, a difference in update frequency emerges among joint angles, and this can cause over-fitting.
  To decrease the difference of update frequency among joint angles, in addition to the data $(\bm{\theta}_{update},\bm{l}_{update})$ which we want to update, we use the data of the initial value $(\bm{0}, \bm{0})$ (meaning when all joint angles are 0, muscle lengths are also 0) and the sets of data which are collected randomly from the current JMM (in this study, 8 sets of data), and update NN by dataset $\bm{D}$ as minibatch.
  \begin{align}
    \bm{D} = \{(\bm{\theta}_{update},\bm{l}_{update}),\;\; (\bm{0}, \bm{0}),\;\; {(\bm{\theta}_{rand}, f(\bm{\theta}_{rand}))}_{1 \cdots N}\}
  \end{align}
  The concept is shown in the upper left figure of \figref{figure:neural-network-system}.
  We succeeded in updating NN online gradually without destroying the entire value of JMM.
}%
{%
  本研究の主眼である実機のデータからの関節-筋空間マップの再学習には, 幾何モデルから得たNNの重みを逐次的に更新していく必要がある.
  しかし, 単純に更新をしてしまうと関節角度によっては更新頻度に差が出てしまうため, その部分だけ過学習をしてしまう.
  そこで, 更新したい値$(\bm{\theta}_{update},\bm{l}_{update})$に加えて、初期値である$(\bm{0}, \bm{0})$(関節角度が全て0のとき、筋長も全て0), 現在のマップからランダムで収集したいくつかのデータ(本研究では8個)を含めた$\bm{D}$をミニバッチとして学習させることによって学習頻度のムラを下げる試みを行った.
  \begin{align}
    \bm{D} = \{(\bm{\theta}_{update},\bm{l}_{update}), (\bm{0}, \bm{0}), {(\bm{\theta}_{rand}, f(\bm{\theta}_{rand}))}_{1..N}\}
  \end{align}
  概念図は\figref{figure:neural-network-system}の左上図のようになっており, この方法によって全体のマップを崩すことなくNNを少しずつ再学習していくことに成功した.
}%

\subsection{The Estimation of Joint Angles}
\switchlanguage%
{%
  Tendon-driven musculoskeletal humanoids do not have sensors measuring joint angles such as encoders and potentiometers because they have many ball joints imitating joints of human beings.
  So, we need to estimate current joint angles from expansion and contraction of muscles.
  In this study, to estimate joint angles, we use the method of Ookubo, et al. \cite{humanoids2015:okubo:muscle-learning}.
  This method uses extended kalman filter (EKF), and we substitute the NN output of this study for the muscle jacobian and the function calculating muscle lengths from joint angles in this method.
  Specifically, for muscle jacobian $G$ and the function $f$ in the state equation and observation equation of EKF as shown below, we use the value obtained by NN output.
  \begin{align}
    \bm{\theta}_{k|k-1} &= \bm{\theta}_{k-1|k-1} + G^{-1}(\bm{\theta})\delta\bm{l} \\
    \bm{l} &= f(\bm{\theta}_{k|k-1})
  \end{align}
}%
{%
  筋骨格ヒューマノイドは人間を模した球関節が多く存在するため, 関節にエンコーダなど, 角度を測るセンサが備わっていない.
  そのため, 筋肉の伸び縮みから現在の関節角を推定する必要がある.
  本研究ではその手法として大久保らの手法\cite{humanoids2015:okubo:muscle-learning}を用いる.
  これはEKF(Extended Kalman Filter)を用いた手法であり, この手法のうちの筋長ヤコビアンと, 関節から筋長への変換部を本研究で構築したNNに置き換える.
  具体的には、EKFにおける以下の状態方程式と観測方程式の$G$, $f$として、NNから得たものを用いる。
  \begin{align}
    \bm{\theta}_{k|k-1} &= \bm{\theta}_{k-1|k-1} + G^{-1}(\bm{\theta})\delta\bm{l} \\
    \bm{l} &= f(\bm{\theta}_{k|k-1})
  \end{align}
}%

\section{Online Learning System} \label{sec:3}
\subsection{Overview}
\begin{figure}[htb]
  \centering
  \includegraphics[width=1.0\columnwidth]{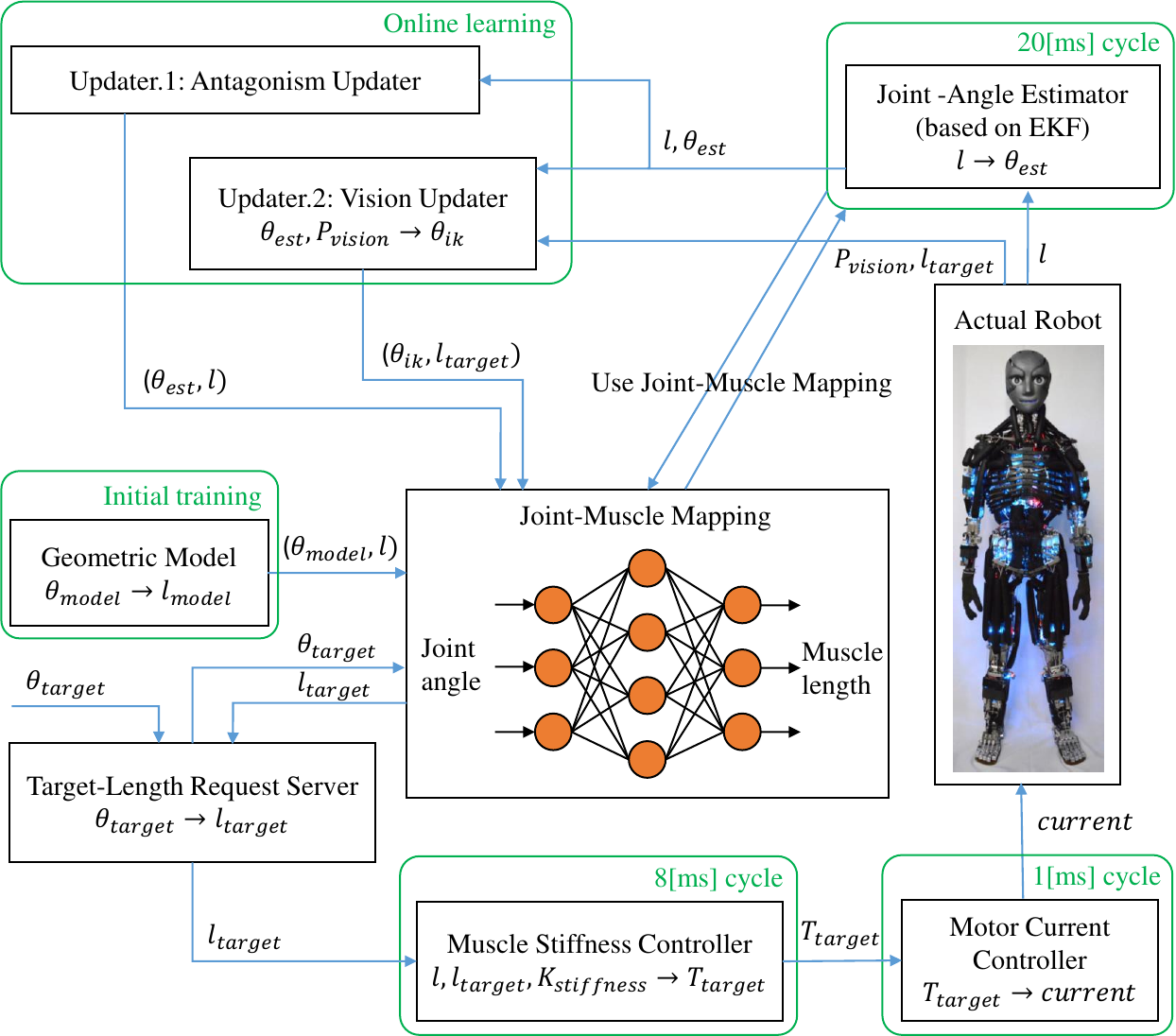}
  \vspace{-3.0ex}
  \caption{System overview.}
  \label{figure:system-overview}
  \vspace{-1.0ex}
\end{figure}

\switchlanguage%
{%
  The overview of this system is shown in \figref{figure:system-overview}.
  We express JMM using NN, and we move the actual robot and estimate joint angles using this JMM.
  Also, we update this JMM online using two types of methods.
  The first method is online learning using estimated joint angles and the actual muscle lengths (Antagonism Updater).
  By this method, we can modify the slack of tendon wires and internal muscle tension between agonist and antagonist muscles that occur due to the model error between the geometric model and actual robot.
  The second method is online learning using the actual joint angles estimated by vision and target muscle lengths sent to move the actual robot (Vision Updater).
  By this method, we can decrease the model error between the geometric model and actual robot, and become able to move the actual robot accurately as intended.
}%
{%
  システム全体の概要図は\figref{figure:system-overview}のようになっている.
  関節-筋空間マップをNNで表し, これを用いてロボットの動作指令, 関節角度の推定を行う.
  また, 二種類の方法によってこの関節-筋空間マップを逐次的に修正していく.
  一つ目の方法は現在の関節角度推定値と実機の筋長を用いた更新であり, これによって幾何モデルと実機の間の誤差による主動筋拮抗筋間における内力の高まりや筋の緩みを修正していく(以後antagonism updaterと呼ぶ).
  二つ目の方法は視覚から推定した現在の関節角度と動作の際に送った筋長を用いた更新であり, これによって物体把持などにおける関節の角度位置指令の誤差をなくし, 指令した通りの場所に身体を動作させることができるように修正する(以後vision updaterと呼ぶ).
}%

\subsection{Antagonism Updater} \label{subsec:antagonism-learning}
\switchlanguage%
{%
  This online learning method is very simple; we update JMM using the method explained at \secref{subsec:relearn} in which the data set is the current estimated joint angles $\bm{\theta}_{est}$ and actual muscle lengths $\bm{l}$ as shown below.
  \begin{align}
    (\bm{\theta}_{update},\bm{l}_{update}) = (\bm{\theta}_{est}, \bm{l})
  \end{align}
  We update JMM when the change of muscle lengths is smaller than a certain value and when $\bm{\theta}_{est}$ is further from previous $\bm{\theta}_{update}$ by more than a certain value.
  We will explain the reason why this online learning is able to modify muscle antagonism correctly.
  First, we move the tendon-driven musculoskeletal humanoid using the procedure below.
  \begin{enumerate}
    \item We set the target joint angles.
    \item We calculate target muscle lengths by inputting the target joint angles into JMM.
    \item We calculate target muscle tensions using muscle stiffness control \cite{robio2011:shirai:control}, and send these to the actual robot.
  \end{enumerate}
  The equation of muscle stiffness control is shown as below.
  \begin{align}
    \bm{T}_{target} = \bm{T}_{bias} + \textrm{max}\{0, \bm{K}_{stiff}(\bm{l}-\bm{l}_{target})\}
  \end{align}
  This control permits muscle length errors to a certain degree, meaning that it can inhibit internal muscle tension and slack of tendons to a certain degree.
  This effect of inhibition becomes large when we make $\bm{K}_{stiff}$ smaller and $\bm{T}_{bias}$ bigger.
  If there are slack of tendons, $\bm{T}_{bias}$ makes the slack decrease, and by using the Antagonism Updater in this situation, the antagonism relationship is modified correctly in the direction that decreases the slack of tendons.
  Also, if there is large internal muscle tension, $\bm{l}$ becomes longer than $\bm{l}_{target}$ due to muscle stiffness control, and the internal muscle tension decreases.
  By using the Antagonism Updater in this situation, we can modify the antagonism relationships correctly in the direction that decreases the large internal muscle tension.
  Thus, if the sent target muscle lengths are impossible to be executed and large internal muscle tension and slack of tendons emerge, this problem can be solved by updating NN using estimated joint angles and actual muscle lengths which in actuality have no slack of tendon wires and smaller internal muscle tension than the original due to muscle stiffness control.
}%
{%
  このオンライン学習の手法は非常にシンプルであり, 以下のように現在の関節角度推定値$\bm{\theta}_{est}$と実機の筋長$\bm{l}$をデータとして, \secref{subsec:relearn}で説明した再学習を行う.
  \begin{align}
    (\bm{\theta}_{update},\bm{l}_{update}) = (\bm{\theta}_{est}, \bm{l})
  \end{align}
  ただし, 筋の長さ変化がある一定より小さくなった瞬間のみ更新を行い, また, 一度更新した関節角度からある程度離れた時のみ更新を行う.
  このオンライン学習手法によって拮抗関係が正しくなる理由は以下のようになっている.
  まず, 筋骨格ヒューマノイドを動作させる際は以下のような順で動作させる.
  \begin{enumerate}
    \item ターゲットとなる関節角度を設定する
    \item 関節-筋空間マップにターゲットとなる関節角度を入れ, ターゲットとなる筋長を算出する
    \item 筋剛性制御\cite{robio2011:shirai:control}を用いてターゲットとなる筋張力を算出し, それを実機に送る
  \end{enumerate}
  ここで, 筋剛性制御の式は以下のようになっており, この制御によって筋長誤差の許容が行われ, 筋の緩みや内力の高まりがある程度抑制される.
  \begin{align}
    \bm{T}_{target} = \bm{T}_{bias} + \textrm{max}\{0, \bm{K}_{stiff}(\bm{l}-\bm{l}_{target})\}
  \end{align}
  この抑制効果は$\bm{K}_{stiff}$を小さく, $\bm{T}_{bias}$を大きくすることでより顕著となる.
  もし実機へ送った筋長に緩みがあった場合, $\bm{T}_{bias}$によって筋の緩みは減り, その際の関節推定値と実機の筋長をオンラインで学習させることによって, その時の関節角度が正しい保証はないが, 緩みが減る方向に拮抗関係自体は正しくなる.
  また, もし大きな内力が発生するような筋長を実機へ送った場合, $\bm{l}$は$\bm{l}_{target}$に比べて少し長くなり, 内力がある程度減った状態に収束する.
  この際の関節推定値と実機の筋長をオンラインで学習させることにより, 内力が減る方向に拮抗関係は正しくなる.
  つまり, 最初は実機に送った筋長に全く誤差がなく一切緩まず内力も小さいということは少ないが, 筋剛性制御によって実際に緩みがなく, ある程度内力が少ない状態になった実機の筋長と関節角度推定値を用いてNNを更新すれば, その拮抗関係の問題は解決されていくのである.
}%

\subsection{Vision Updater} \label{subsec:vision-learning}
\switchlanguage%
{%
  In this section, we will explain the method to modify JMM using the RGB camera of Kengoro and AR marker attached to the hand of Kengoro.
  This method can be applied to not only the hand but also any other part of the body, and we can use the feature value of the body or the value of IMU instead of the AR marker.
  The procedure of this online learning is stated as below.
  \begin{enumerate}
    \item By looking at the AR marker attached to the hand from the RGB camera in the head, we can obtain the relative position and orientation from the camera to the hand.
    \item We obtain the actual joint angles $\bm{\theta}_{ik}$ by solving inverse kinematics (IK), in which the initial joint angles are the current estimated joint angles, the target position and orientation are $\bm{P}_{vision}$ as stated above, and the links are the head, thorax, collarbone, scapula, humerus, ulna and radius.
      \begin{align}
        \bm{\theta}_{ik} = IK(\bm{\theta}_{initial}=\bm{\theta}_{est}, \bm{P}_{target}=\bm{P}_{vision})
      \end{align}
    \item Having omitted any mistakes of the IK (omitted if the actual joint angles $\bm{\theta}_{ik}$ are too different from the current estimated joint angles $\bm{\theta}_{est}$), we execute online learning using the sent target muscle lengths and actual joint angles.
      \begin{align}
        (\bm{\theta}_{update},\bm{l}_{update}) = (\bm{\theta}_{ik}, \bm{l}_{target}) \;\;\;\;\; if \; ||\bm{\theta}_{ik} - \bm{\theta}_{est}||_2 < C
      \end{align}
      where $||\cdot||_2$ expresses L2 norm and $C$ is a threshold.
  \end{enumerate}
  Like the Antagonism Updater, we update JMM when the change of muscle lengths is smaller than a certain value and when $\bm{\theta}_{ik}$ is further from previous $\bm{\theta}_{update}$ by more than a certain value.

  This method is different from the Antagonism Updater in that this updater can modify the JMM to become able to move the actual robot accurately as intended.
  We can use the actual muscle lengths $\bm{l}$ instead of the target muscle lengths $\bm{l}_{target}$, in which case we are also able to modify JMM accurately to a certain degree.
  However, due to muscle stiffness control, there is a difference between the sent muscle lengths and actual muscle lengths.
  So when using $\bm{l}$, it is difficult to meet requirements needed to move the actual robot to the intended position, and using $\bm{l}_{target}$ is better regarding tracking accuracy.
}%
{%
  このオンライン学習の手法の手順は以下のようになっている.
  ここでは, 腱悟郎についたRGBカメラから自身の手のARマーカを見ることによって関節-筋空間マップを修正する方法を説明するが, この手法は手だけでなく, 足や腹など, 様々な部位に対して適用でき, また, ARマーカではなく手の特徴量などを用いることも可能である.
  \begin{enumerate}
    \item 手についたARマーカをRGBカメラで見ることで, 自身の目から自身の手までの相対位置姿勢を得る
    \item 現在の関節角度推定値を初期値, 上記の相対位置姿勢$\bm{P}_{vision}$をターゲットとし, 目から首, 胸郭, 鎖骨, 肩甲骨, 上腕, 前腕をリンクとした逆運動学(IK)を解いてその際の関節角$\bm{\theta}_{ik}$を得る.
      \begin{align}
        \bm{\theta}_{ik} = IK(\bm{\theta}_{initial}=\bm{\theta}_{est}, \bm{P}_{target}=\bm{P}_{vision})
      \end{align}
    \item その関節角度が関節角度推定値と極端に大きく異なっていなければ得られた関節角度を実機の関節角度とし, これとターゲットとして送った筋長を用いてオンライン学習を実行する
      \begin{align}
        (\bm{\theta}_{update},\bm{l}_{update}) = (\bm{\theta}_{ik}, \bm{l}_{target}) \;\;\;\;\; if \; ||\bm{\theta}_{ik} - \bm{\theta}_{est}||_2 < C
      \end{align}
  \end{enumerate}
  ここで、$||\cdot||_2$はL2ノルムを表す。
  ただし, 前節と同様に筋の長さ変化がある一定より小さくなった瞬間のみ更新を行い、また、一度更新した関節角度からある程度離れた時のみ更新を行う.
  この手法は前節とは違い, 絶対的な関節角度自体を正しくすることができる.
  ここではターゲットとなる筋長$\bm{l}_{target}$ではなく、実機の筋長$\bm{l}$を用いることも可能であり、その場合もある程度は関節-筋空間マッピングが修正される。
  しかし、筋剛性制御の影響によって送った筋長と実際の筋長は異なり、$\bm{l}$を用いる場合はある位置まで手を動かしたいという要求を満たすことは難しく、$\bm{l}_{target}$を用いた場合の方が追従精度は良くなる。
}%

\section{Experiments} \label{sec:4}
\switchlanguage%
{%
  We conducted some experiments using the proposed system.
  First, we will discuss the construction of JMM using NN, and next, we will verify the two types of online learnings, respectively.
  After that, we will integrate the two online learnings, and execute quantitative analysis on the realization of intended joint angles.
  Finally, through a manipulation experiment, we will verify the effectiveness of this proposed system.
}%
{%
  本システムを用いたいくつかの実験を行う。
  まずはニューラルネットワークによる関節-筋空間マップの構築に関する議論を行い、次に、二種類の再学習方法を一つずつ検証する。
  その後、それらを統合した学習方法を用いたオンライン学習について、意図した関節角度の実現性について定量的な実験を行う。
  最後に、物体把持実験を行うことで、本手法の有効性を示す。
}%

\begin{figure}[htb]
  \centering
  \includegraphics[width=1.0\columnwidth]{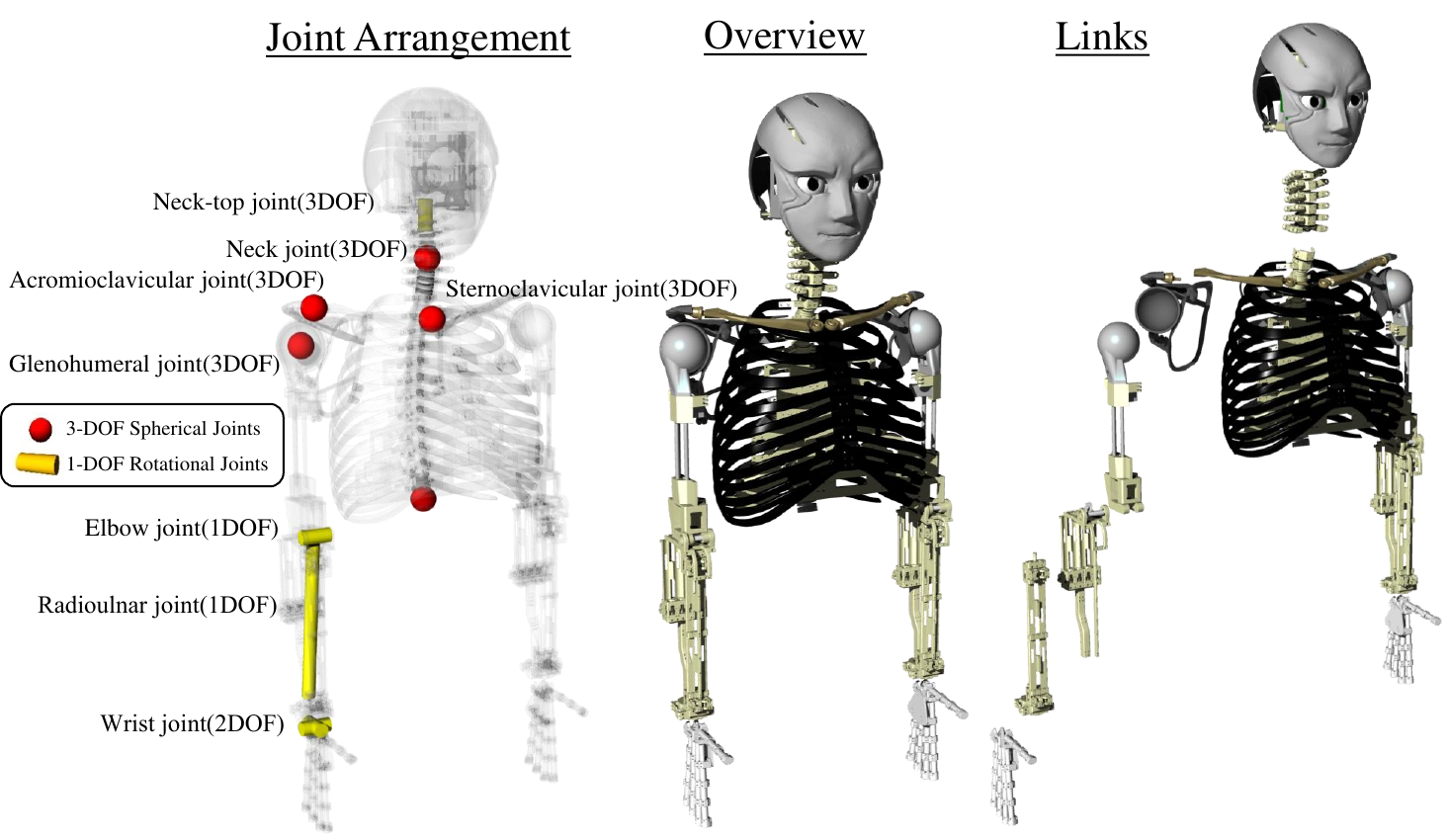}
  \caption{Joint structure of Kengoro upper limb.}
  \label{figure:kengoro-link}
\end{figure}

\begin{figure}[htb]
  \centering
  \includegraphics[width=1.0\columnwidth]{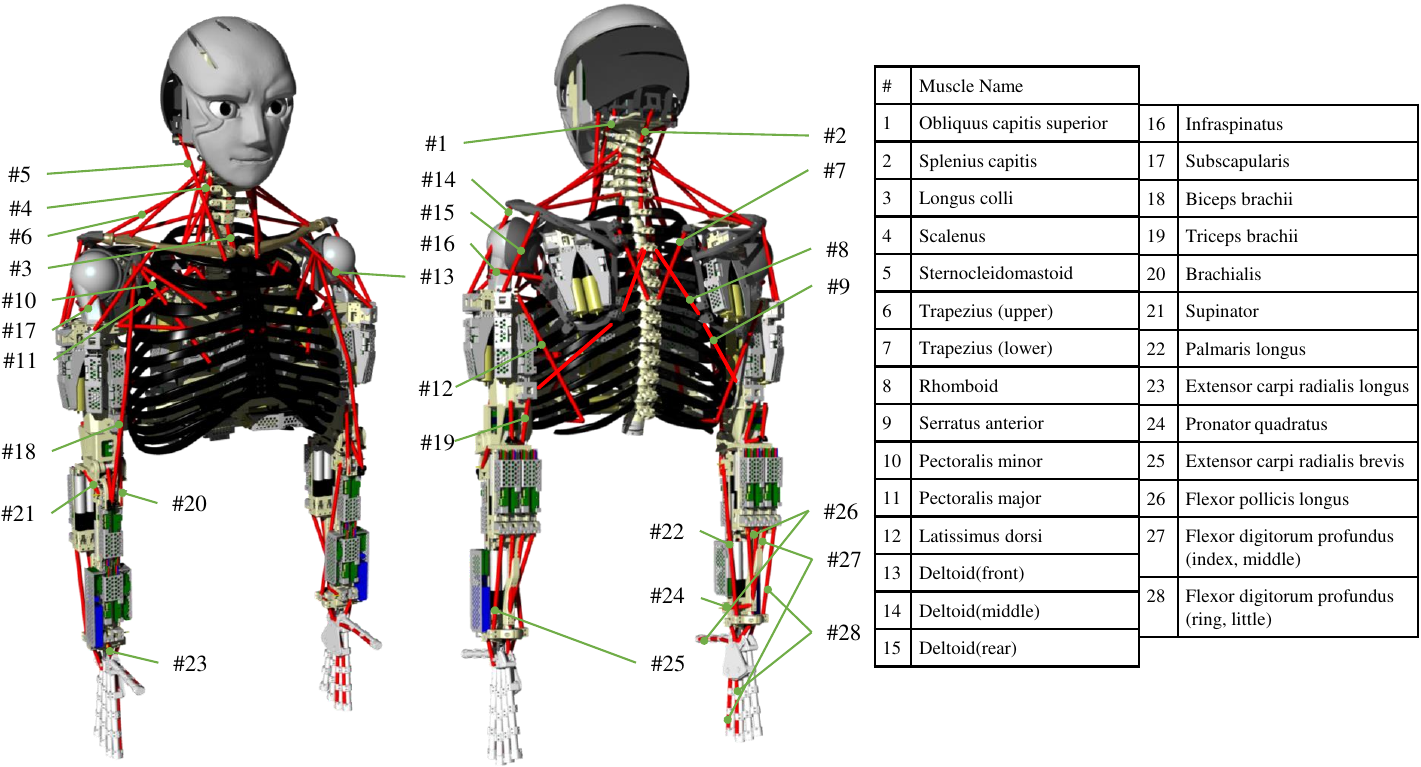}
  \caption{Muscle arrangement of Kengoro upper limb.}
  \label{figure:kengoro-muscle}
\end{figure}

\subsection{Basic Experiments of Joint-Muscle Mapping} \label{subsec:basic-experiments}
\begin{figure}[htb]
  \centering
  \includegraphics[width=1.0\columnwidth]{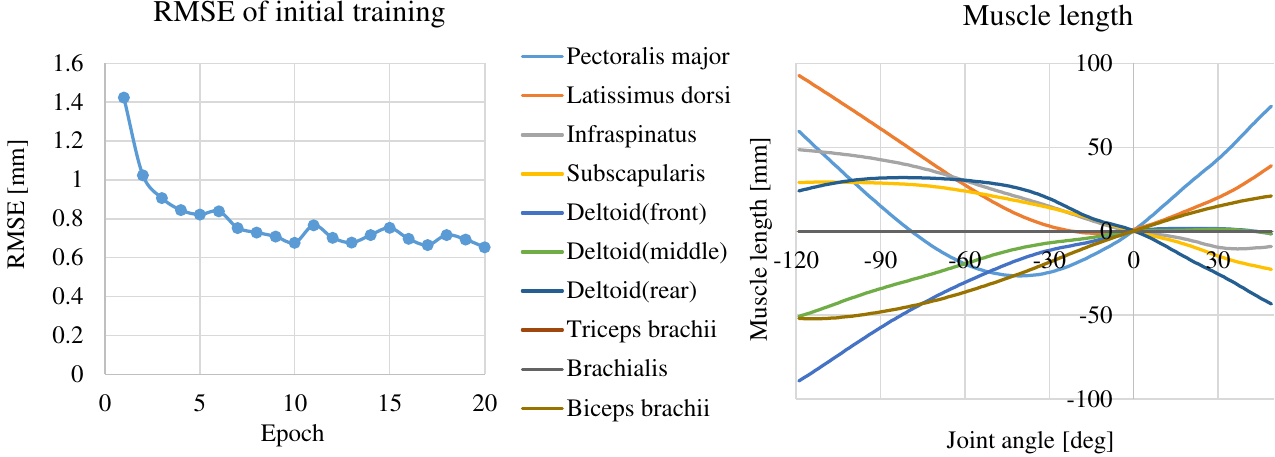}
  \caption{The result of the initial training experiment. Left graph shows the loss transition of initial training using the geometric model; right graph shows example of muscle length change during shoulder flexion.}
  \label{figure:neural-network-experiment}
\end{figure}

\begin{figure}[htb]
  \centering
  \includegraphics[width=1.0\columnwidth]{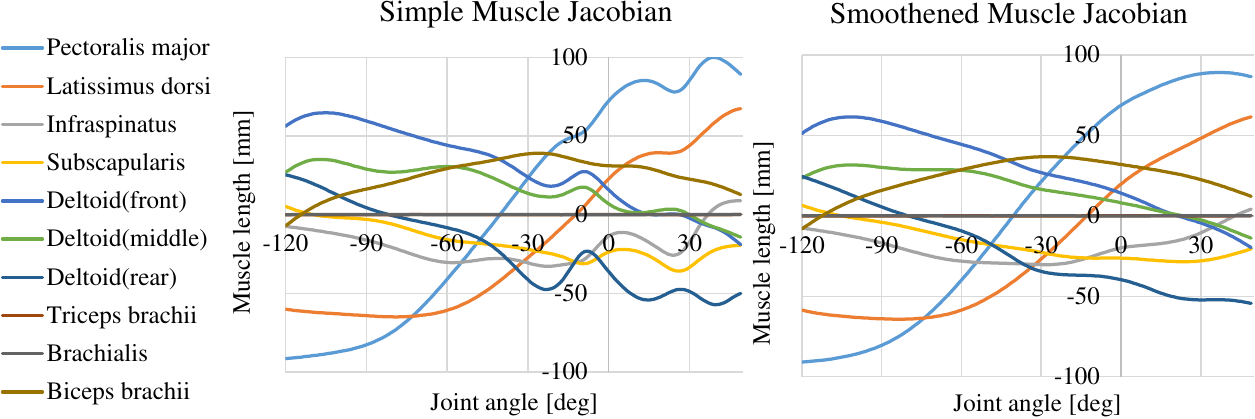}
  \caption{Muscle jacobian in the pitch direction during shoulder flexion. Left graph is the analytical differentiation of NN; right graph is the differentiation of NN using the method of \secref{subsec:jacobian}.}
  \label{figure:neural-network-jacobian}
\end{figure}

\switchlanguage%
{%
  First, we will consider the initial training of JMM from a geometric model.
  The joint structure of Kengoro's upper limb (\figref{figure:kengoro-link}) is the same as in human beings, and we divided it into 4 groups: the neck, scapula, shoulder, and forearm.
  As stated in \secref{subsec:initial-learning}, we constructed 4 JMMs using NN regarding these respective groups.
  In this study, the number of hidden layer units is 1000, and the activation function of NN is sigmoid.
  In the preliminary experiment considering the structure of NN, in the case of 8 DOFs and 10 muscles included in the shoulder JMM, which is the biggest in this study, we verified that one hidden layer and 1000 units in the hidden layer are enough to express the JMM (loss of training is under 1 [mm] as shown in \figref{figure:neural-network-experiment}).
  The unit of input joint angles is [rad], and the unit of output muscle lengths is [mm].
  As an example, we show the muscle arrangement of Kengoro upper limb in \figref{figure:kengoro-muscle} and the JMM of the shoulder of Kengoro in \figref{figure:neural-network-experiment}.
  The data set of joint angles and muscle lengths is obtained by equally dividing the range of each joint movement into 5--9 parts and moving the joint angles of the geometric model by each value, and the number of total data was 2646000.
  In this initial training, the size of minibatch is 5, the number of epoch is 20, the optimization method is Adam, and we used one-fifth of the data set randomly as validation.
  The loss transition, or the root mean squared error (RMSE), when training is shown in the left graph of \figref{figure:neural-network-experiment}.
  As an example, we show the change of 10 muscle lengths from NN when moving the shoulder pitch axis in the right graph of \figref{figure:neural-network-experiment}.
  From these graphs, we can see that the weight of NN is trained well to a value similar to that of the geometric model.

  Second, regarding muscle jacobian, we show the muscle jacobian in the pitch direction during shoulder flexion in \figref{figure:neural-network-jacobian}.
  In other words, it is a differentiation of the right graph of \figref{figure:neural-network-experiment}.
  When we differentiated simply using analytical differentiation of NN, the graph fluctuated as shown in the left graph of \figref{figure:neural-network-jacobian}.
  In comparison, when we differentiated using the method of \secref{subsec:jacobian}, a smooth and natural graph was obtained as shown in \figref{figure:neural-network-jacobian}.
}%
{%
  腱悟郎の上肢の関節構造は人間と同様な構造をしており, これを首・肩甲骨・肩・前腕の4つに分けて考える.
  \secref{subsec:initial-learning}と同様にそれぞれのパートを分け, これら一つ一つに関節-筋空間マップのNNを構築する.
  本研究では全ての隠れ層のユニット数を1000, 活性化関数をsigmoidとしている.
  NNの構造については, 実験から関節-筋空間マップを表すのに隠れ層は一層で十分であり, 本研究で最も大きい関節数8, 筋数10に対して必要なユニット数は1000あれば足りる(例えば\figref{figure:neural-network-system}の左図に示すように肩ではRMSEが1.0[mm]以下と小さい)ということを予備実験で確認した.
  入力する関節角度はradian単位であり, 出力する筋長はmm単位である.
  例として肩に関係する筋の関節-筋空間マップを\figref{figure:neural-network-experiment}に示す.
  データは幾何モデルにおいて各関節の最小角度から最大角度までを6--9等分してそれぞれ動作させた際の関節角度と筋長の組であり, データ数は2646000個である.
  本研究では比較的詳細に幾何モデルからデータを取ったためデータ数が多いが, 実際にはランダムに関節角度データを100000個ほど取ることでも学習は可能である.
  今回のminibatchのサイズは5, epoch数は20, 最適化手法はAdamであり, この内毎回ランダムに1/5を検定に使っている.
  学習時の検定の二乗平均平方根(RMSE)の推移は\figref{figure:neural-network-experiment}の左図のようになっており, 例として肩のpitch軸を動作させた際の10本の筋に関するNNでの筋長変化は\figref{figure:neural-network-experiment}の右図のようになっている.
  これらグラフからわかるように幾何モデル通りに関節-筋空間マップの重みが学習できていることがわかる.

  また、筋長ヤコビアンに関してだが、\figref{figure:neural-network-experiment}の右図のときの関節角度での肩のpitch軸方向に対する筋長ヤコビアン、つまり、右図の微分を求めてみる。
  単純に0.1[deg]の差分を取る方法で微分をした場合、\figref{figure:neural-network-jacobian}の左図に示すように、波打った値となってします。
  それに対して、\secref{subsec:jacobian}の手法で微分をした場合、右図のように変動の小さなより自然な筋長ヤコビアンが得られていることがわかる。
}%

\subsection{Experiment of Antagonism Updater using Elbow Joint}
\begin{figure}[htb]
  \centering
  \includegraphics[width=1.0\columnwidth]{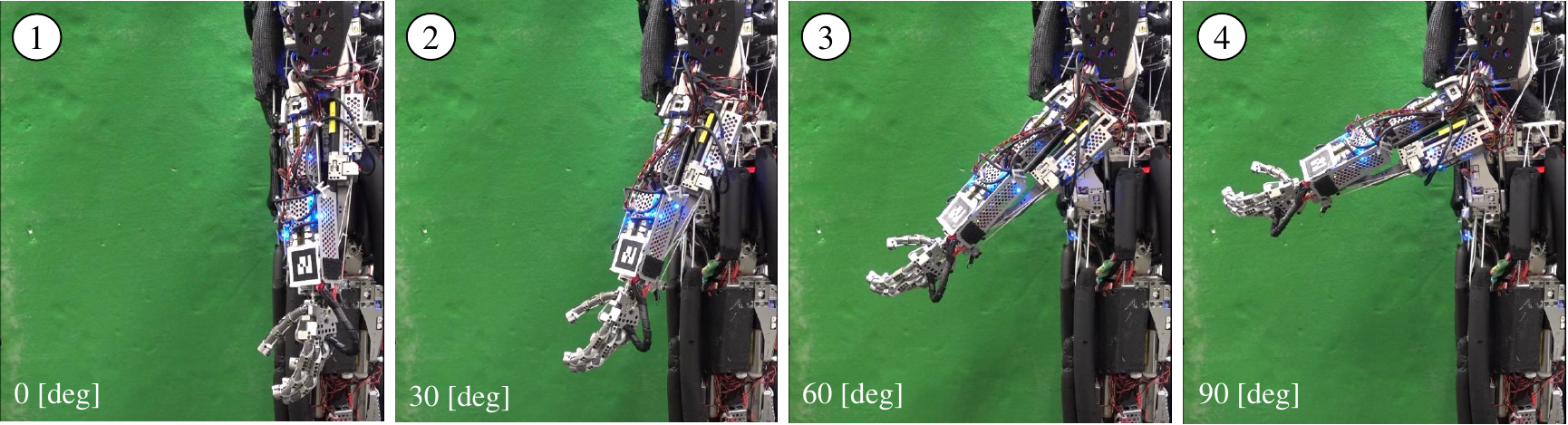}
  \caption{The movement of Antagonism Updater experiment using the elbow joint. We repeated this movement 11 times.}
  \label{figure:elbow-antagonistic-update-video}
\end{figure}

\begin{figure}[htb]
  \centering
  \includegraphics[width=1.0\columnwidth]{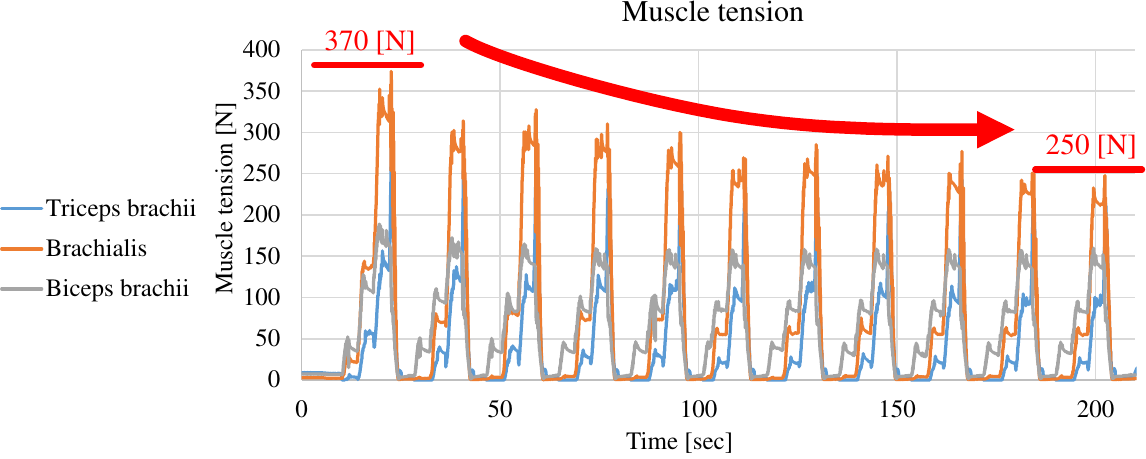}
  \caption{The result of Antagonism Updater experiment using the elbow joint. This graph shows muscle tension change during elbow flexion.}
  \label{figure:elbow-antagonistic-update}
\end{figure}

\switchlanguage%
{%
  We will show the effectiveness of Antagonism Updater stated in \secref{subsec:antagonism-learning} through an experiment using the elbow joint.
  In the elbow joint, there are mainly 3 muscles: the brachialis, biceps brachii, and triceps brachii.
  In this experiment, we moved the elbow joint of Kengoro up to 90 [deg] by 30 [deg] in 2 second intervals (\figref{figure:elbow-antagonistic-update-video}) continuously.
  We show the muscle tensions during this experiment in \figref{figure:elbow-antagonistic-update}.
  The graph shows that the two agonist muscle tensions became equal and the antagonist muscle tension decreased gradually during the movement of the elbow joint.
  In this experiment, maximum muscle tension decreased from 370 [N] to 250 [N] in 11 trials.
  Although there was a large difference in muscle tensions among agonist muscles (the brachialis and biceps brachii) at first, the difference decreased gradually during the trials.
  By this experiment, we could verify the effectiveness of the Antagonism Updater.
  However, as shown in \figref{figure:elbow-antagonistic-update-video}, the joint angles are largely different from the intended angles.
  This updater only modifies the muscle antagonism.
  In order to move the actual robot as intended, we conducted an experiment of Vision Updater next.
}%
{%
  \secref{subsec:antagonism-learning}で述べた関節角度推定値と実機の筋長を用いたオンライン学習の有効性を, 肘関節を用いた実験で示す.
  肘関節には主に3つの筋が存在し, それは上腕筋, 上腕二頭筋, 上腕三頭筋である.
  動作としては, 肘を0度, 30度, 60度, 90度の順に2秒で動かして2秒止めるを繰り返すような動作を行う(\figref{figure:elbow-antagonistic-update-video}).
  肘関節を何度も動かすことによって, 主動筋二本の筋張力が均等になりかつ拮抗筋の筋張力が減っていく様子を\figref{figure:elbow-antagonistic-update}に示す.
  11回の試行で最大張力は37[kgf]から25[kgf]まで減っている。
  最初は発揮張力に大きな差があった主動筋である上腕筋と上腕二頭筋の張力であるが、施行ごとに段々と近くなっているのがわかる。
  これによって本オンライン学習の有効性が示された.
  しかし、\figref{figure:elbow-antagonistic-update-video}に見るように、実際に動作させたい関節角度と実機の動作が大きく異なっていることがわかる。
  このUpdaterは拮抗関係の修正のみを行い、実機が指令した通りの動作を行うようになるために、次節ではVision-Updaterに関する実験を行う。
}%

\subsection{Experiment of Vision Updater}
\begin{figure}[htb]
  \centering
  \includegraphics[width=1.0\columnwidth]{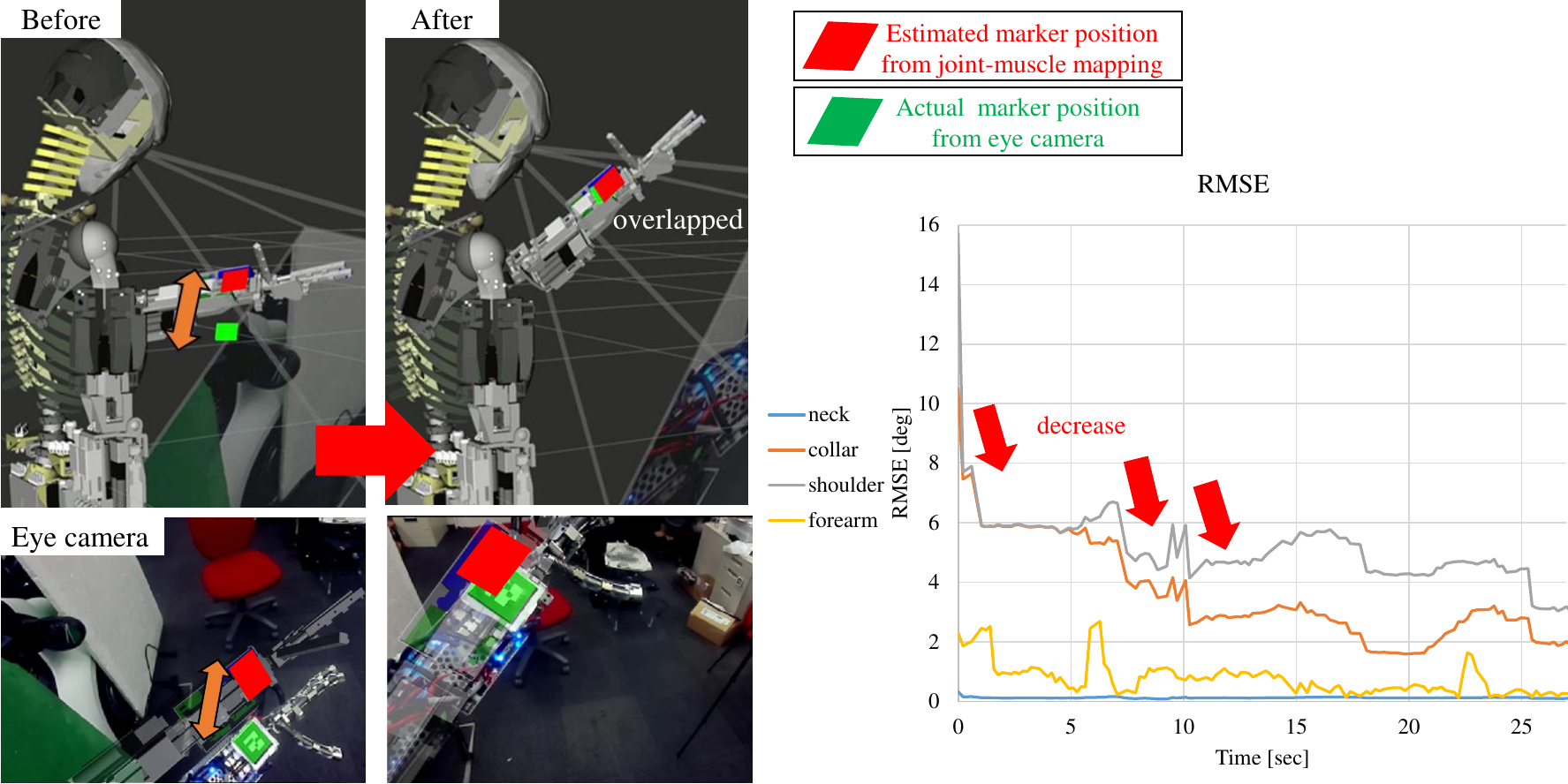}
  \caption{The result of Vision Updater experiment. Left figure shows how this experiment was carried out; right graph shows RMSE of the difference between the estimated joint angles from JMM and the actual joint angles using the RGB camera.}
  \label{figure:vision-update}
\end{figure}

\switchlanguage%
{%
  In order to verify the effectiveness of Vision Updater, we will show the correct modification of estimated joint angles during an experiment moving the upper limb of Kengoro and looking at the limb.
  In this experiment, we sent various joint angles to Kengoro's upper limb, Kengoro looked at the AR marker attached to the hand from eyes in the head, and updated JMM online using the information.
  The correct modification of JMM by this updater is shown in \figref{figure:vision-update}.
  The left figure shows how the estimated position of the AR marker from the current JMM and the position of the AR marker from Kengoro's RGB camera became close using this updater.
  Although there was a difference among the two AR marker positions at first, the positions became closer and finally overlapped.
  The right graph shows the transition of RMSE of the difference between the estimated joint angles from JMM and the actual joint angles estimated using vision, and the RMSE became smaller gradually.
  For example, the RMSE of the shoulder was about 16 [deg] at first, but it finally decreased to about 3 [deg].
  By this experiment, we verified the effectiveness of Vision Updater.
}%
{%
  \secref{subsec:vision-learning}で述べた視覚とターゲットとなる筋長を用いたオンライン学習の有効性を示すため, 腕を動作させ, その際の体を見ることで関節角度推定が正しくなっていく様子を示す.
  人間が腱悟郎の首や肩に任意の姿勢を連続的に送り, その際に頭に搭載したカメラによって手についたマーカを認識し, それを用いてオンライン学習を行う.
  それによって関節角度推定が正しくなる様子を\figref{figure:vision-update}に示す.
  左図は手についたマーカの関節-筋空間マッピングからの推定位置と、目から見たマーカの位置が正しくなっていく様子を示している.
  最初は二つのマーカの一がズレているが、最終的には重なるほど近い位置を示すようになっている。
  また, 右図は関節角度推定の値と目で見たマーカからIKを解いた際の関節角度の値の差のRMSEの推移を表しており, その値が小さくなっていく様子を示している.
  shoulderを例に取ると、最初はRMSEが約16[deg]だったのに対して、最終的に約3[deg]程度まで下がっている。
  これらによって, 視覚を用いたオンライン学習が有効であることが示された.
}%

\begin{figure}[htb]
  \centering
  \includegraphics[width=1.0\columnwidth]{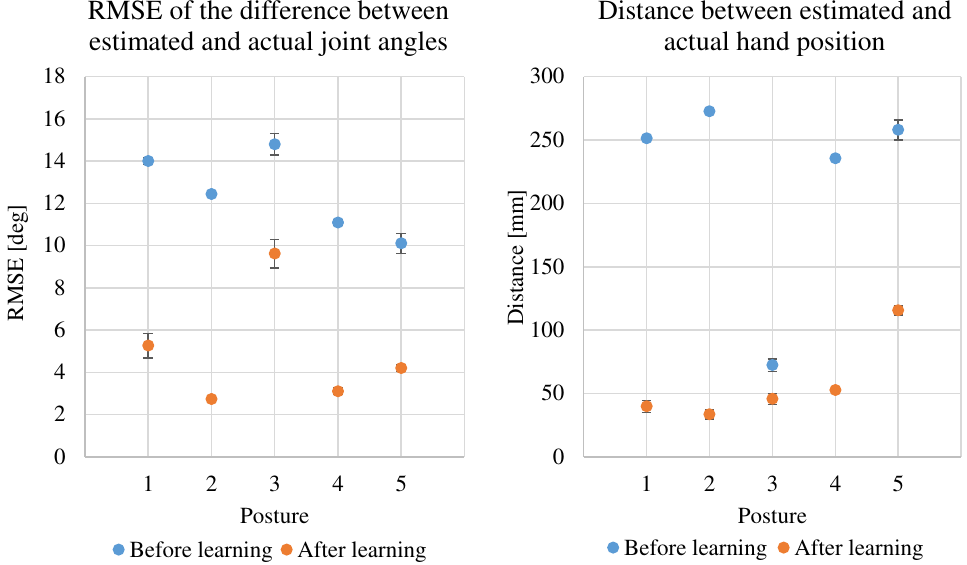}
  \caption{RMSE of the difference between the estimated and actual joint angles $RMSE_{joint}$ and the distance between the estimated and actual hand xyz position $Distance_{hand}$, regarding before and after the online learning using Antagonism Updater and Vision Updater. We showed the average and standard deviation among several trials. The error bar of this graph is the standard deviation.}
  \label{figure:ave-dev}
\end{figure}

\begin{figure*}[t]
  \centering
  \includegraphics[width=2.0\columnwidth]{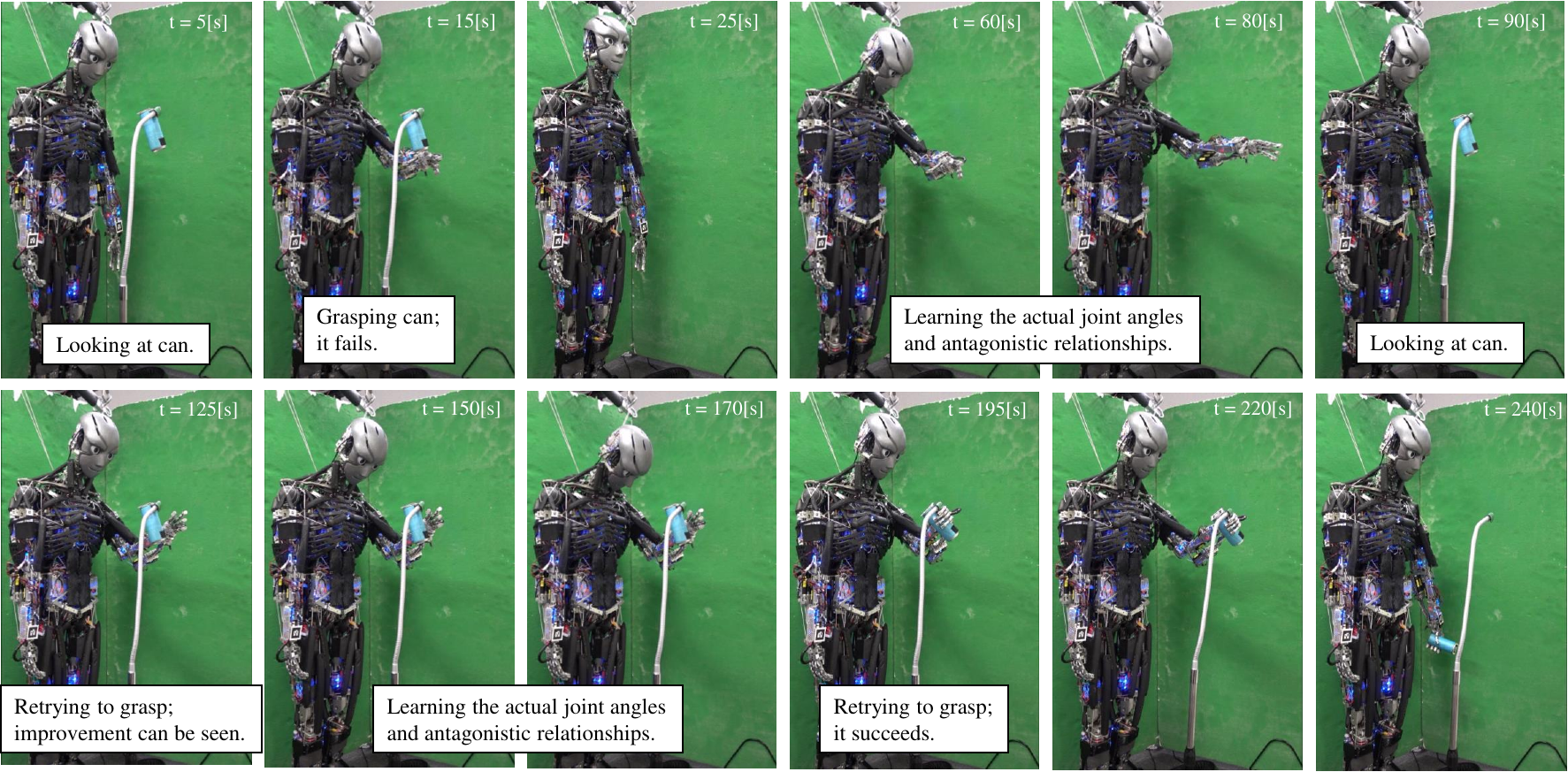}
  \caption{The result of can grasping experiment. This figure shows how the experiment was carried out.}
  \label{figure:cylinder-test}
\end{figure*}

\subsection{Experiment of Antagonism Updater and Vision Updater}
\switchlanguage%
{%
  We will integrate the two updaters: Antagonism Updater and Vision Updater, apply them to the actual robot, and show quantitative analysis regarding realization of intended joint angles.
  We evaluated the average and standard deviation of RMSE of the difference between the estimated and actual joint angles (RMSE of ($\bm{\theta}_{est}-\bm{\theta}_{actual}$), $RMSE_{joint}$) by repeating the movement from randomized start postures to target postures.
  As a reference, $\bm{\theta}_{est}$ is almost equal to the $\bm{\theta}_{target}$ we sent, because $\bm{\theta}_{est}$ is estimated using muscle lengths $\bm{l} = f(\bm{\theta}_{target})$.
  First, we sent several postures to the robot successively, conducted online learning using the actual robot, and obtained modified JMM.
  The number of postures we sent during online learning was 27 in 300 [sec].
  Next, we generated 5 target postures (named posture 1--5; we checked these postures are feasible and absolute error of joint angles between these postures and all postures we sent during the online learning are distant by at least 20 [deg]), and repeated the movement from 10 randomized start postures to postures 1--5, respectively.
  We moved the robot from randomized start postures in order to consider the influence of hysteresis and reproducibility.
  We conducted these trials before and after the online learning (total: 100 trials), and showed $RMSE_{joint}$ in \figref{figure:ave-dev}.
  There are some differences among target postures 1--5, but the average of $RMSE_{joint}$ decreased from before to after the online learning regarding all postures.
  The standard deviation is very small.
  Also, we showed the distance between the estimated and actual hand xyz position $Distance_{hand}$ in \figref{figure:ave-dev}.
  We can see $Distance_{hand}$ decreases regarding all target postures 1--5.
  In total, the average of $RMSE_{joint}$ among all trials decreased from 12.49 [deg] to 4.99 [deg]: about 40\%, and that of $Distance_{hand}$ decreased from 217.95 [mm] to 57.53 [mm]: about 26\%, regarding before and after the online learning.
  Thus, the two types of online learning do not compete, have generalization ability, and work effectively.
}%
{%
  二つのupdaterであるantagonism updaterとvision updaterを統合して実機に適用し、その際の意図した関節角度の実現性に関して定量的な評価を与える。
  ランダムに定義した姿勢に、ランダムに生成した姿勢から遷移することを繰り返し、意図した関節角度がどの程度実現されるかを、意図した関節角度と実際の関節角度との差分のRMSEの平均分散を得ることで評価する。
  まず、二つのupdaterを使用して実機においてオンライン学習を行った。
  人間が事前に決めた任意の姿勢をロボットに連続的に送り、その際にオンライン学習を実行させる。
  この際に得られたVision Updaterにおけるデータはn個、Antagonism Updaterにおけるデータはm個だった。
  その後、ランダムに姿勢を5つ生成し(これをposture 1,2,3,4,5と名付ける)、それぞれの点に対して、ランダムに生成した姿勢からその姿勢に遷移させること10回ずつ繰り返した。
  ランダムに生成した姿勢からある姿勢に変化させたのは、ヒステリシス等の影響や再現性を見るためである。
  これをonline learningを行う前と後に関してそれぞれ行い(計100試行)、それぞれの姿勢になる際の、意図した姿勢と実際の姿勢の差分のRMSEの平均と分散を\figref{figure:ave-dev}に示す。
  ターゲットとなる姿勢によって多少の違いはあるものの、online learningをする前と後で意図した姿勢の実現性が上昇しているのがわかる。
  分散は非常に小さく収まっており、これらonline learningが競合せず、有効に働くことが示せた。

}%

\subsection{Experiment of Can Grasping}
\begin{figure}[htb]
  \centering
  \includegraphics[width=1.0\columnwidth]{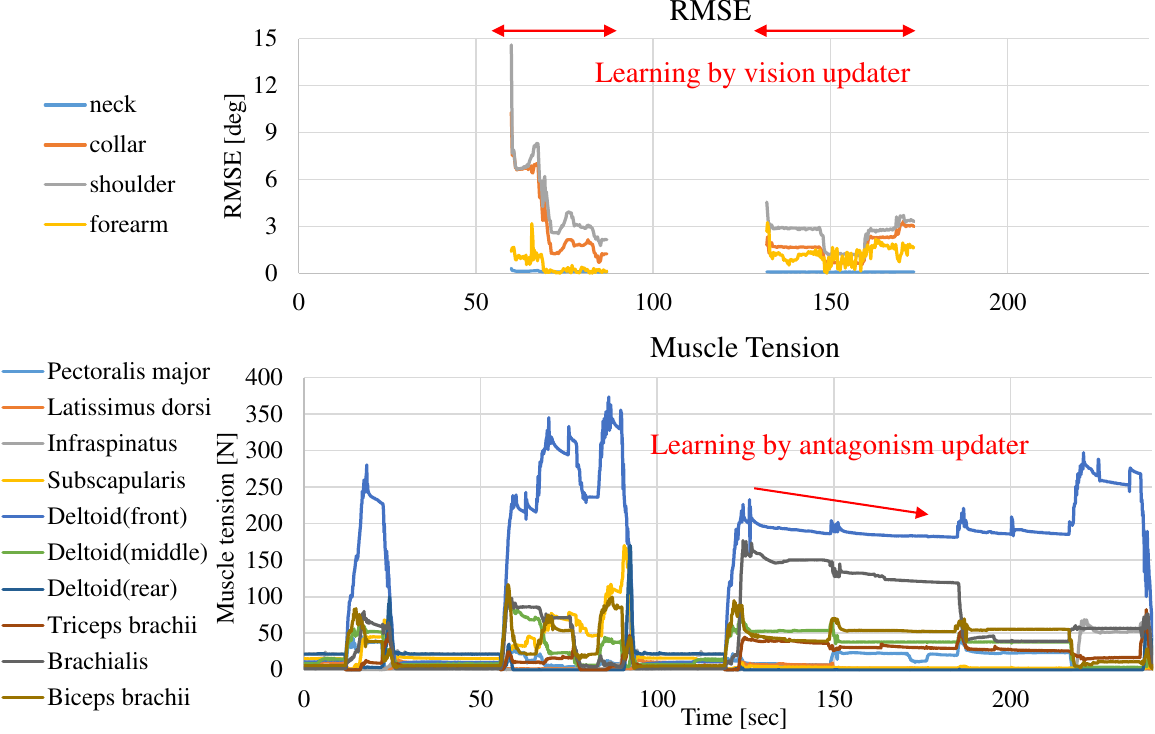}
  \caption{The result of can grasping experiment. Upper graph shows RMSE of the difference between the estimated joint angles and the actual joint angles using the RGB camera; Lower graph shows the muscle tensions during this experiment.}
  \label{figure:cylinder-test-graph}
\end{figure}

\begin{figure}[htb]
  \centering
  \includegraphics[width=1.0\columnwidth]{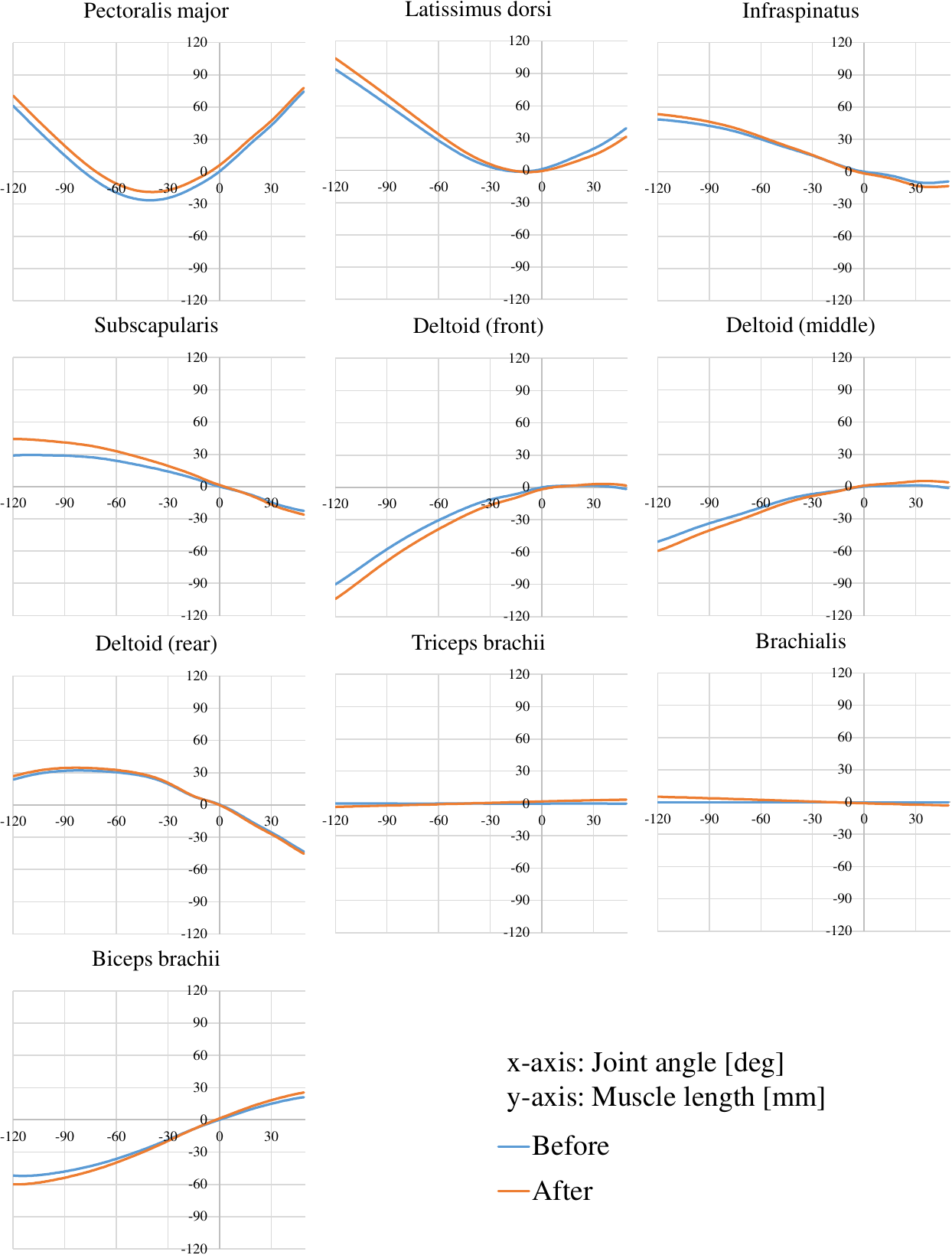}
  \caption{The result of can grasping experiment. These figures compare the transition of the 10 muscle lengths in the shoulder during shoulder flexion between before the update of JMM and after.}
  \label{figure:cylinder-test-jmm}
\end{figure}

\switchlanguage%
{%
  We conducted a manipulation experiment of looking at a can and grasping it, by integrating the entire system proposed in this study.
  In all experimental movements, the proposed two updaters are executed.
  How this experiment was carried out is shown in \figref{figure:cylinder-test}.
  Also, RMSE of the difference between the estimated joint angles and the actual joint angles using the RGB camera and muscle tensions during this experiment are shown in \figref{figure:cylinder-test-graph}.
  We will explain the detailed movements of this experiment.
  First, Kengoro looked at a can with an AR marker, solved IK at the position of the AR marker, and moved the hand to the position in order to grasp it.
  However, at first, there were some errors between the geometric model and actual robot, and Kengoro could not move the hand to the intended position.
  Second, Kengoro moved the upper limb to various positions, and looked at the hand.
  During these movements, JMMs that are accurate to a certain degree were obtained using Vision Updater.
  Then, Kengoro approached the can again.
  The approach to the can became better, but Kengoro could not approach it completely, so he looked at the upper limb again and updated JMM online.
  Finally Kengoro was able to approach the can correctly and grasp it.
  In this experiment, the positions at which IK is solved are completely the same, but each actual position of the moved hand is different every time, because the JMM is modified correctly during the movement.
  Also, as shown in \figref{figure:cylinder-test-graph}, the RMSE and the required muscle tensions decreased gradually.
  Finally, in order to visualize the difference of the JMM before and after its update, as an example, we show the transition of 10 muscle lengths in the shoulder during shoulder flexion before and after the update in \figref{figure:cylinder-test-jmm}.
  Although the overall shape is not destroyed, we can see a difference of 15 [mm] at -120 [deg] between before the update and after.

  By this experiment, we verified that this study is effective.
}%
{%
  本研究で提案した全システムを統合し, 缶を見てそれを握らせるような物体把持実験を行う.
  全ての動作中に, 提案した二種類のオンライン学習が動作している.
  その際の実験の様子を\figref{figure:cylinder-test}に示す.
  また、実験中のRMSEと筋張力については\figref{figure:cylinder-test-graph}に示す。
  実験の詳細な動作を説明する.
  まず, ARマーカのついた缶を見て, それに対してIKを解き, 缶を掴む動作を行う.
  しかし, 最初は幾何モデルと実機の間の誤差により正しい位置まで腕は動作せず, 掴むことはできない.
  次に, 様々に体を動かしてみる.
  この間に視覚を用いた逐次的な学習によってある程度正しい関節-筋空間マップが得られている.
  そして, もう一度缶に対してアプローチを行う.
  缶に対するアプローチは良くなったものの、まだ届いていないため、最後にもう一度周りを見渡して逐次的再学習を行う。
  そして最終的に缶を掴み取ることができている。
  この際、缶に対してIKを解いている位置は常に同じであるが、関節-筋空間マッピングが正しく修正されていっているため、手が動作する場所は変わっていっているのがわかる。
  また、\figref{figure:cylinder-test-graph}に示すように、RMSEが段々と減り、必要な筋張力が減っているのがわかる。
  最後に、関節-筋空間マッピングの修正を行う前と修正を行った後における関節-筋空間マッピングの違いを可視化するために、例として、shoulder-flexionの際の肩の10本の筋の筋長の変化を修正前と修正後について、\figref{figure:cylinder-test-jmm}に示す。
  全体の形は崩れないまま、修正前と修正後で-120[deg]のときに最大で15[mm]程度まで違いが現れているのがわかる。

  この実験によって, 二種類のオンライン学習が競合しないこと, そして本システムの有効性が示された.
}%

\section{CONCLUSION} \label{sec:5}
\switchlanguage%
{%
  In this study, we showed the method of online learning of joint-muscle mapping (JMM) which modifies the JMM accurately using vision and the actual robot information, in tendon-driven musculoskeletal humanoids which have large model error between the geometric model and actual robot.
  Although JMM is usually expressed by polynomial regression, table-searching, and so on, we expressed JMM by neural network (NN), allowing for the online update of JMM.
  We discussed the method of initial training from a geometric model, online learning that prevents over-fitting, and obtaining smooth and natural muscle jacobian.
  Also, we developed two online updaters of JMM.
  First, online learning using the current estimated joint angles and actual muscle lengths can modify muscle antagonism correctly, preventing large internal muscle tension and slack of tendon wires (Antagonism Updater).
  Second, online learning using the actual joint angles estimated from vision and sent target muscle lengths can modify the JMM so that the robot can move to the intended position (Vision Updater).
  By these online learning methods of JMM, the error between the target and actual joint angles decrease to 40\% in 5 minutes, and a manipulation task by tendon-driven musculoskeletal humanoids which have not been done until now due to the model error between the geometric model and actual robot becomes possible.
  This study is one step for tendon-driven musculoskeletal humanoids to approach and exceed ordinary humanoids driven by actuators in each axis.

  In future works, we would like to try two tasks.
  First, although in this study, we used AR markers attached to the hand in order to obtain the hand position and orientation, the robot should recognize the movement of the hand itself and obtain the information about position and orientation.
  Second, we want to add the effect of muscle tension to JMM, and manipulate objects more accurately while changing muscle stiffness.
  In this study, we were unable to consider muscle elongation due to muscle compliance, so next, we should implement online learning of a new model $\bm{l} = f(\bm{\theta}, \bm{T})$ which updates not only joint-muscle mapping, but also the influence of muscle compliance.
}%
{%
  本研究では, 実機と幾何モデル間の誤差が非常に大きい筋骨格ヒューマノイドにおいて, 視覚や実機の筋長情報を用いることで関節-筋空間マップが正しくなっていくことを示した.
  関節-筋空間マップは今まで多項式近似などで表現されることが多かったのに対して, それをNNによって表すことで逐次的なマップの更新を可能とした.
  NNにおける過学習を防ぐようなオンライン学習方法、滑らかな筋長ヤコビアンの導出などに関して議論を行った。
  また, この関節-筋空間マップの更新則を二つ開発した.
  一つ目として, 現在の関節角度推定値と実機の筋長を用いたオンライン学習によって, 筋の緩みや内力の高まりを回避するような拮抗関係の学習を行うことができることがわかった.
  二つ目として, 視覚と関節角度推定値から現在の正しい関節角度を推定し, それを用いることで位置誤差を修正していくような学習を行うことができることがわかった.
  これら関節-筋骨格マッピングの逐次的な再学習によって、今まで実機-幾何モデル間の誤差ゆえに行うことのできなかった筋骨格ヒューマノイドにおけるマニピュレーション動作が可能となった。

  今後は, 二つの課題に取り組みたい.
  一つ目は, 現在はARマーカを用いて手の位置を取得しているのに対して, それをロボット自身が自分の動きから手を自分のものだと認識し, 位置などの情報を抽出するということである.
  二つ目は、この関節-筋空間マッピングに筋張力の影響を考慮し、剛性を操作しながらの精巧なマニピュレーションを行いたい。
  本研究では、外力が加わっていないような状況を考えたため、筋のコンプライアンスによる伸び等を考慮していなかった。
  今後は、筋張力の影響を加えた$\bm{l} = f(\bm{\theta}, \bm{T})$というモデルの学習方法についても考えていきたい。
}%

{
  \bibliographystyle{IEEEtran}
  \bibliography{main}
}

\end{document}